\documentclass{article}
\usepackage{ijcai23}

\usepackage{times}
\usepackage{soul}
\usepackage{url}
\usepackage[hidelinks]{hyperref}
\usepackage[utf8]{inputenc}
\usepackage[small]{caption}
\usepackage{graphicx}
\usepackage{amsmath}
\usepackage{amsthm}
\usepackage{booktabs}
\usepackage{algorithm}
\usepackage{algorithmic}
\usepackage[switch]{lineno}

\usepackage{xspace}

\author{
Vineel Nagisetty$^1$\and
Laura Graves$^1$\and
Guanting Pan$^1$\And
Piyush Jha$^1$\And
Vijay Ganesh$^1$
\affiliations
$^1$University of Waterloo
\emails
\{vineel.nagisetty, laura.graves, g6pan, piyush.jha, vganesh\}@uwaterloo.ca
}

\title{CGDTest: A Constrained Gradient Descent Algorithm for\\ Testing Neural Networks}

\begin{document}

\newcommand{\toolname}{CGDTest\xspace}
\newcommand{\numoftool}{32\xspace}

\maketitle

\begin{abstract}
In this paper, we propose a new Deep Neural Network (DNN) testing algorithm called the Constrained Gradient Descent (CGD) method, and an implementation we call CGDTest aimed at exposing security and robustness issues such as adversarial robustness and bias in DNNs. Our CGD algorithm is a gradient-descent (GD) method, with the twist that the user can also specify logical properties that characterize the kinds of inputs that the user may want. This functionality sets CGDTest apart from other similar DNN testing tools since it allows users to specify logical constraints to test DNNs not only for $\ell_p$ ball-based adversarial robustness but, more importantly, includes richer properties such as disguised and flow adversarial constraints, as well as adversarial robustness in the NLP domain. We showcase the utility and power of CGDTest via extensive experimentation in the context of vision and NLP domains, comparing against \numoftool state-of-the-art methods over these diverse domains. Our results indicate that CGDTest outperforms state-of-the-art testing tools for $\ell_p$ ball-based adversarial robustness, and is significantly superior in testing for other adversarial robustness, with improvements in PAR2 scores of over 1500\% in some cases over the next best tool. Our evaluation shows that our CGD method outperforms competing methods we compared against in terms of expressibility (i.e., a rich constraint language and concomitant tool support to express a wide variety of properties), scalability (i.e., can be applied to very large real-world models with up to 138 million parameters), and generality (i.e., can be used to test a plethora of model architectures).
\end{abstract}

\section{Introduction} \label{sec:introduction}

Over the last decade, Deep Neural Networks (DNNs) have become ubiquitous and are being used in various settings from image recognition to safety-critical systems~\cite{sarker2021machine}. Simultaneously and unsurprisingly, we are witnessing a variety of attacks against DNNs, including adversarial, data-poisoning, membership inference, and model inversion attacks, to name just a few~\cite{huang2020survey}. More broadly, there is a clear recognition in the machine learning community that the problem of lack of reliability and security of ML models - in particular DNNs - is of concern and needs to be addressed effectively~\cite{VNNLIB}. 

Recognizing the aforementioned problem, researchers have proposed a variety of testing, analysis, and verification (TAV) methods that have been specifically tailored to the DNN setting~\cite{huang2020survey}. \textbf{Testing} methods aim to find inputs that satisfy some constraints often using a variant of the gradient-descent (GD) method~\cite{huang2020survey}. While testing methods scale well and are effective for a variety of model architectures, they do not provide any guarantees of correctness of models with respect to security or robustness properties. Further, they often do not allow users to specify logical properties, and are tailored narrowly to some well-known set of properties such as $\ell_p$ ball-based adversarial robustness. By contrast, \textbf{Verification/program analysis} methods symbolically ``capture" the model and often use a solver to either find inputs that violate given properties or guarantee that the DNN adheres to said properties~\cite{katz2019marabou}. Unfortunately, verification methods suffer from lack of scalability and generality (i.e., can only handle certain kinds of architectures), but tend to be very expressive. Finally, \textbf{reachability} methods define an allowable input region and propagate that region through the network, resulting in an output domain that encapsulates all possible outputs given any input described in the input region~\cite{singh2019abstract}. These methods perform better than verification techniques on scalability and generality, but not as well as testing. (A detailed overview of each type of method is provided in the Related Work Section~\ref{sec:related})

In many industrial settings, model developers demand testing/verification methods that can scale to millions of parameters (scalability), are suitable for many architectures (generality), and are expressive enough to allow users to specify a variety of properties of interest (expressibility). Hence, the focus of our work is to design a practical DNN testing tool that scores well on scalability, generality, and expressibility. 

\noindent{\bf Problem Statement:} More precisely, the research question we address is the following: is it possible to create a testing method that allows model developers to specify properties (without requiring the DNN-under-test be converted into a slow-to-verify symbolic form), while at the same time be able to scale to real-world models and be general enough to handle a wide variety of model architectures?

\vspace{0.1cm}
\noindent {\bf Overview of CGD Algorithm:} In order to address the above research question, we propose a method called the Constrained Gradient Descent (CGD) that enhances gradient descent (GD) methods with the ability to specify logical constraints to test DNNs against specific properties that are not currently supported by most of the commonly used testing, verification, or reachability tools. Specifically, CGD converts a rich class of logical constraints into a differentiable loss function (called `constraint loss') following the approach introduced in Probabilistic Soft Logic~\cite{psl}. This constraint loss can be minimized using a GD algorithm. 

\subsection{Our Contributions:}

\begin{enumerate}
    \item \textbf{Constrained Gradient Descent (CGD) Algorithm and its Implementation:}  Our key contribution is the constrained gradient descent algorithm (and a tool that we call \toolname). This method allows users to specify logical constraints as part of the loss function of a GD algorithm, with the goal of creating DNN test inputs that satisfy these user-specified logical constraints. Our method addresses the most important concerns of model developers: 1) a testing tool that scales to industrial-strength DNNs ({\textbf scalability}), 2) is architecture-neutral ({\textbf generality}), and most importantly, 3) enables users to logically specify the kinds of inputs they want ({\textbf expressibility}). The class of constraints expressible via CGD is larger than other gradient-based methods, allowing a CGD-based approach to be applied in contexts that have more complex input constraints. (See Section~\ref{sec:cgd}.)

    \item \textbf{Showcasing the Scalability, Generality, and Expressibility of \toolname on NLP Domain:} To showcase that \toolname is capable of testing for richer classes of properties than $\ell_p$ ball-based adversarial properties in the vision domain, we pick the domain of Natural Language Processing (NLP) for adversarial robustness. Current DNN TAV tools built for the vision domain do not work well in the domain of NLP because of the discrete nature of the input space, as well as inherent semantic and syntactic constraints~\cite{textattack}. We introduce a method to apply gradient-based testing techniques in the domain of NLP and then compare it against 5 state-of-the-art adversarial attack tools designed for the NLP domain on a target model trained on a semantic classification dataset. We show that the \toolname is more effective and efficient than competing SOTA tools in producing adversarial examples. (See Section~\ref{sec:results}).
    
    \item \textbf{Showcasing the Scalability, Generality, and Expressibility of \toolname on Vision Domain:} We report on detailed experiments where we empirically evaluate the scalability and generality of \toolname and compare it with state-of-the-art testing, verification, and reachability methods over multiple benchmark suites consisting of models with sizes ranging from 6000 to {\textbf 138 million parameters}. We show that using the $\ell_p$ ball-based definition of adversarial examples, the \toolname tool is comparable to state-of-the-art DNN TAV tools. Further, in the disguised and flow adversarial examples, \toolname showed massive improvements while other methods scored much lower, either because they couldn't scale or didn't allow properties to be specified or placed restrictions on the DNN architecture. (See Section~\ref{sec:results}.)    
\end{enumerate}

\section{Background} \label{sec:background}

\subsection{Adversarial Examples} \label{sec:adversarial}
Adversarial examples, for a given DNN-under-attack, are inputs that are similar (under a suitable mathematical {\textit similarity metric}) to a correctly classified input, and at the same time cause the DNN-under-attack to classify them incorrectly. Often, such similarity metrics are defined using $\ell_p$ norms over the input space. An example $x'$, under an $\ell_p$-norm similarity metric with threshold $\epsilon$, is said to be adversarial if the following holds:

\begin{center}
    $(\|x-x'\|_p \leq \epsilon) \wedge (argmax(M(x)) \neq argmax(M(x')))$
\end{center}

Researchers have developed various testing and verification methods for finding adversarial examples~\cite{huang2018survey}. There are also annual competitions, such as the VNN-COMP~\cite{vnncomp2021report}, that are aimed at evaluating DNN testing (aka, incomplete solvers) and verification (aka, complete solvers) tools against a benchmark of models and adversarial robustness properties. 

\subsection{Individual Fairness} \label{sec:fairness}
Dwork et al. summarized individual fairness as ensuring that ``similar individuals are treated similarly"~\cite{fairness}. In other words, any two individuals that are similar w.r.t. a certain task should be classified similarly. Formally, given a (often complex) distance function $d$ that captures similarity between two individuals $x$ and $x'$, a model $M$ can be considered unfair using this constraint if:
\begin{center}
    $(d(x,x') \leq \epsilon) \wedge (argmax(M(x)) \neq argmax(M(x')))$
\end{center}
In practice, selecting a good distance function can be challenging, and may require a subject expert's opinion~\cite{fairness}. Alternatively, as shown in~\cite{individual_fairness_old}, this distance can be learned from data. In this work, the distance is computed as: $d_{x}^{2}(x,x') = \langle x-x',P(x-x')\rangle$, where $P$ defines a projection matrix orthogonal to some sensitive subspace. 

\subsection{Converting Constraints to Loss} \label{sec:constraintLoss}
The language of constraints in our setting consists of boolean combinations of comparisons between terms, where each term is either a constant or a differentiable function that is defined over inputs, intermediate layer values, and outputs of the DNN. In order to convert the constraint $\varphi$ to a differentiable loss function, we perform a simple parsing, where we map boolean connectives as well as comparisons using \textit{Lukasiewicz t-norm} and its corresponding \textit{co-norm}. This translation is designed to be exact at the extremes (`completely true' or `completely false') and provides a consistent mapping for values in between, allowing us to calculate smooth gradient information. Examples of these relaxations are provided below. 

\begin{enumerate}
        \item $L(\varphi' \wedge \varphi'') := L(\varphi') + L(\varphi''),$
        \item $L(\varphi' \vee \varphi'') := L(\varphi') * L(\varphi'')$.
\end{enumerate}

This results in a function that is differentiable and continuous (almost) everywhere. We refer to this function as `constraint loss'. A comprehensive list detailing the terminology used as well as more information on the constraint translation process is provided in Appendix A. 

\section{The CGD Algorithm} \label{sec:cgd} \label{sec:cgd_alg}

In this section, we provide an overview of our constrained gradient descent (CGD) method (refer Algorithm~\ref{alg:cgd}) which uses gradient descent over the constraint loss described in Section~\ref{sec:constraintLoss}. The input to the CGD algorithm is a DNN-under-test $M$, a label $y$, a weight $\alpha$, and a constraint $\varphi$ that encodes some DNN property of interest. The output of the algorithm is a DNN input $x$ that satisfies the constraints $\varphi$ and $(argmax(M(x))=y)$. 

\begin{algorithm} [t]
\caption{The CGD Algorithm}
\label{alg:cgd}
\begin{algorithmic}[1]
  \REQUIRE DNN $M$, Label $y$, Weight $\alpha$, Constraint $\varphi$
  \ENSURE DNN Input $x$ s.t. $(argmax(M(x))=y) \wedge \varphi(x)$ 
  \STATE $\varphi'$ = ConstraintLoss($\varphi$)
  \STATE initialize $x$
  \WHILE{True}
    \IF{$(argmax(M(x))=y)$ AND $\varphi'(x)=0$}
        \STATE break
    \ENDIF
    \STATE $l_{1}$ = $\varphi'(x)$ //compute constraint loss
    \STATE $l_{2}$ = Loss($M(x),y$)
    \STATE $\nabla x_1$ = $\textstyle \frac{\partial l_1}{\partial x}$ 
    \STATE $\nabla x_2$ = $\textstyle \frac{\partial l_2}{\partial x}$ 
    \STATE $x$ = $x - \alpha * (\nabla x_1 + \nabla x_2)$
  \ENDWHILE
  \RETURN $x$
\end{algorithmic}
\end{algorithm}

In the CGD algorithm, we first parse user-specified constraints $\varphi$ and convert them into a constraint loss $\varphi'$ function via a short piece of code (line 1 - also refer to Appendix~\ref{app:conversion}). The constraint loss function is constructed in such a way that it evaluates to 0 only when all constraints are satisfied. We initialize the input $x$ to the DNN $M$ as a random vector, whose size is equal to the size of the input of $M$ (line 2). We then enter a loop that terminates when the constraints are satisfied (i.e., $\varphi'(x)=0$), and the model classifies $x$ as the intended label $y$ (lines 4-6). If $x$ satisfies these criteria, we break out of the loop and return it as the solution (line 13). Otherwise, we calculate both constraint loss (line 7) and classification loss (line 8) and then calculate the gradient of both losses with respect to $x$ (lines 9 and 10). Finally, we combine both gradient values and alter $x$ slightly (using a small $\alpha$ value) to move in the direction of decreased loss (line 11). This process continues until either $x$ satisfies both criteria or the tool reaches a timeout.

Intuitively, we can think of the gradients for constraint and misclassification loss (lines 9 and 10 in Algorithm~\ref{alg:cgd}) as having distinct loss landscapes traversed by the CGD algorithm to find a global minima. If the loss landscapes do not intersect, or a point that acts as global minima to both landscapes does not exist, there is no possible solution. However, even if such a point exists, CGD is not guaranteed to find a solution, consistent with other GD algorithms. This is because, one or both losses may be non-convex, and so this algorithm may get stuck in a local minima. 

It is for this reason that we add a small amount of carefully crafted noise (similar to~\cite{yu2022gradient}) to help avoid local minima and allow us to find results faster. We may also wish to find inputs that `mostly' satisfy the constraints. In this case, we could add a small bound $\delta$ and update our conditional statement in line 4 to check whether the constraints are mostly satisfied (i.e., $\varphi'(x) < \delta$). Further, CGD does not exhaustively search the entire space, so we add a limit to the number of iterations (or amount of time) it is run and return ``incomplete" when it doesn't find a solution within that limit. We may find better results if using separate weights $\alpha_1, \alpha_2$ for each of the gradient values. This may allow us to control the effect of each of the gradient values in order to better alter $x$. Moreover, we also initialize a batch of inputs instead of just one and also make sure that the randomized input abides by the input constraints during this initialization step.

Note that this gradient descent approach has some similarities to the Fast Gradient Sign Method (FGSM)~\cite{goodfellow2014explaining}. By aligning the sign of the perturbation with the sign of the gradient, FGSM searches for adversarial examples by moving in the direction of increased loss using the definition of adversarial examples. However, FGSM does not take constraints into account, and it can only test for adversarial robustness in the vision domain. By contrast, CGD searches for inputs that satisfy constraints by moving towards decreased combined misclassification and constraint loss. Further, CGDTest can also construct inputs for NLP and tabular domains, in addition to the vision domain. 

\subsection {\bf \toolname Tool Implementation Details.} \label{sec:implementation}

We implemented \toolname tool using Tensorflow 2.2. \toolname takes in as input the constraints given in the language as described in Section~\ref{sec:background}. These constraints are then converted to constraint loss using a simple parser as described in Section~\ref{sec:background}. This constraint loss is then input, along with the model $M$ and label $y$, to \toolname in order to find examples that satisfy the constraints. Since CGD can potentially be stuck in an infinite loop, we may terminate \toolname if it reaches a maximum time limit returning ``incomplete". 

\section{Evaluation of the \toolname Tool} 
\label{sec:comparision}
\subsection{Experimental Setup: Hardware and Software} All our experiments were performed on a four Intel i5-4300U (CPU @1.9 GHz) core machine with 16 GB RAM running Ubuntu 20.04. In order to run the experiments, we implemented an evaluation framework using Python 3.8, Tensorflow and Keras and compared state-of-the-art DNN TAV tools on the vision and NLP. Refer to Section~\ref{sec:experiment_adversarial} for details on testing for adversarial robustness in the vision domain, Section~\ref{sec:experiment_nlp} for details on testing for the adversarial robustness on the NLP domain, and Section~\ref{sec:vnn_res} for details on testing against tools and benchmarks from VNN-COMP 2021 and VNN-COMP 2022. 


\begin{table}[t]
    \centering
    \begin{tabular}{|l|r|r|r|}
    \hline
    & \textbf{$\ell_p$ ball} & \textbf{Disguised} & \textbf{Flow} \\ 
    \textbf{TAV Tools} & \textbf{PAR2 (sec)} & \textbf{PAR2 (sec)} & \textbf{PAR2 (sec)} \\ \hline 
    {\bf \toolname} & 18,476.36 & \textbf{143,010.88} & \textbf{45,170.87}\\ \hline
    AutoAttack & \textbf{17,459.21} & * & *\\ \hline
    B\&B & 18,259.62 & * & *\\ \hline
    C\&W & 19,713.19 & * & *\\ \hline
    BIM & 23,120.65 & * & *\\ \hline
    FGSM & 43,220.72 & * & *\\ \hline
    Crown-IBP & 585,099.22 & 606,399.84 & *\\ \hline
    ERAN & 600,040.70 & 647,970.45 & * \\ \hline
    nnenum & 604,635.81 & 642,475.34 & * \\ \hline
    MIP Verify & 653,736.16 & 703,006.80 & *\\ \hline
    DLFuzz & 707,066.27 & * & * \\ \hline
    Marabou & 703,057.30 & 728,376.78 & *\\ \hline
    Fuzz & 734,821.58 & 780,000.00 & 780,000.00 \\ \hline
    Genetic & 703,211.67 & 735,509.32 & 761,186.62\\ \hline
    \end{tabular}
    \vspace{0.1cm}
    \caption{Results of experiments on the VNN-LIB benchmarks from the vision domain, aggregated over all benchmark suites. A * indicates that the tool was not able to test for the given type of adversarial example.}
    \label{tab:all_results}
\end{table}

\noindent {\bf Measuring Results:} In order to compare the various tools, we use attack success rate, as well as PAR2 score, a metric that is commonly used by the SAT/SMT/VNNCOMP solver communities in order to effectively compare various solvers on a suite of benchmarks. PAR2 is calculated as follows: if a tool produces a result (input violating the logical specifications or adherence of property) within a given time limit, we record the time taken (in seconds) to produce that result. Conversely, if a tool is unable to produce a result in the given time, we add a penalty of 2*time-limit. A lower PAR2 score is considered better, as that signifies that the tool was able to find violating inputs (or certify the adherence of the property by DNN-under-test) faster on average over a method that resulted in a higher PAR2 score. 

\subsection{Benchmarks}
\label{benchmarks}
The benchmarks used are:
\begin{enumerate}
    \item {\bf VNN-LIB and VNN-COMP 2021/2022}: The first benchmark suite consists of models from VNN-LIB and VNNCOMP 2021/2022, a series of international competitions aimed at comparing SOTA DNN verification and testing tools~\cite{VNNLIB}. A detailed description of the VNN-LIB benchmark suite can be found in \cite{guidotti2020verification}, and of the VNNCOMP 2021 and 2022 can be found in~\cite{vnncomp2021report,vnncomp2022report}. 
    
    
    
    \item {\bf TweetEval}: This suite consists of a single semantic classification model using the last layers of the BERT model. This model is fine-tuned using the TweetEval dataset, a collection of tweets with the task of classifying whether a given tweet is offensive. While the entire BERT model has over 110 million parameters, our model uses around 100,000 parameters (See Section~\ref{sec:experiment_nlp}). 
\end{enumerate}

\subsection{Adversarial Robustness over the Vision Domain (VNN-LIB)}
\label{sec:experiment_adversarial}

After an extensive and thorough survey of various DNN TAV tools on the vision domain taken from VNN-LIB benchmarks~\cite{VNNLIB}, we chose \textbf{Fast Gradient Sign Method (FGSM)}~\cite{goodfellow2014explaining}, \textbf{Basic Iterative Method (BIM)}~\cite{kurakin2016adversarial}, \textbf{Carlini \& Wagner Attack (C\&W)}~\cite{carlini2017towards}, \textbf{Brendel \& Bethge Attack (B\&B)}~\cite{bandb}, and \textbf{AutoAttack}~\cite{autoattack} from the category of GD methods, \textbf{DLFuzz}~\cite{fuzzy} and \textbf{Random Fuzz} from the category of fuzzing based methods, and \textbf{Genetic Algorithm (GA)} from the category of evolutionary algorithm based methods. From the available verification/program analysis-based methods, we selected \textbf{Marabou}~\cite{katz2019marabou}, \textbf{Crown-IBP}~\cite{crownibp}, \textbf{nnenum}~\cite{bak2021nnenum}, and \textbf{MIP Verify}~\cite{tjeng2017evaluating} tools. From the reachability-based methods, we selected \textbf{ERAN}~\cite{singh2019abstract}. 

Most of these tools were selected due to their impressive performance at the VNN-COMP 2020 contest~\cite{VNNLIB} or for completeness of evaluation (e.g., genetic algorithm). (In Section~\ref{sec:vnn_res}, we include an additional comparison of \toolname against tools and benchmarks from VNN-COMP 2021 and VNN-COMP 2022.)

For our experimental evaluation, we chose a set of benchmark suites consisting of a combination of DNN models locally trained and sourced from the VNN-COMP website~\cite{VNNLIB}. More details about the benchmarks can be found in Section~\ref{benchmarks}. In order to compare the expressibility of these tools, we used constraints to define three types of adversarial examples (refer to Appendix~\ref{app:constraints} for more information):  
\begin{enumerate}
        \item \textbf{$\ell_p$ ball-based Adversarial Constraint}: this constraint uses the $\ell_p$ ball-based definition of adversarial example, using $L_{\infty}$ distance with an $\epsilon$ value of 0.2. 
        \item \textbf{Disguised Adversarial Constraint}~\footnote{Examples to illustrate the relevance of such a constraint is given in Appendix \ref{app:disguised}}: Unlike $\ell_p$ ball-based adversarial constraint, disguised adversarial constraints require the original label $y$ to be the second highest label of $M$. Here, $x'$ is, according to some notion, still similar to $x$ as $M$ has fairly high confidence in $y$ (since it is the second-highest label). In the previous definition using the $\ell_p$ ball-based adversarial constraint, we require the new label on $x'$ be anything \textit{but} $y$, while here we require $y$ to be the \textit{second highest} label with no additional $\ell_p$ ball-based constraint. The reason these examples are interesting is that they characterize a very large class of examples that are similar to a given input but are subtly different and would be classified in top-2 (For example, a disguised malevolent actor trying to fool a face recognition system).
        
       \item \textbf{Flow Adversarial Constraint:} this class of constraints utilizes spatial transformations (flow) as given in \cite{xiao2018spatially} to find adversarial examples. More specifically, given a DNN model $M$, label $y$, original input $x$, and new input $x'$, these constraints check whether $argmax(M(x'))\neq y$, while $x'$ and $x$ have minimal local distortion induced by the flow equation (Equation 1 in~\cite{xiao2018spatially}). Such adversarial constraints are interesting when spatial transformations can be used to deceive a DNN (e.g., an image of a roadblock can be adversarial at certain angles, in that it doesn't look like a roadblock to a DNN-under-attack).

\end{enumerate}

We created a framework that provides the following as input to each tool in our evaluation: a DNN model $M$, an input $x$ to $M$, and a logical specification $\varphi$, and checks whether the tool produces a result within an allotted time of 600 seconds. For the inputs, we randomly selected 25 examples from the test sets of the datasets the target model was trained on. Our framework was set to run all tools using these triples, and record the number of successful adversarial examples generated by each tool. 

\begin{figure}[t]
    \centering
    \includegraphics[width=\linewidth]{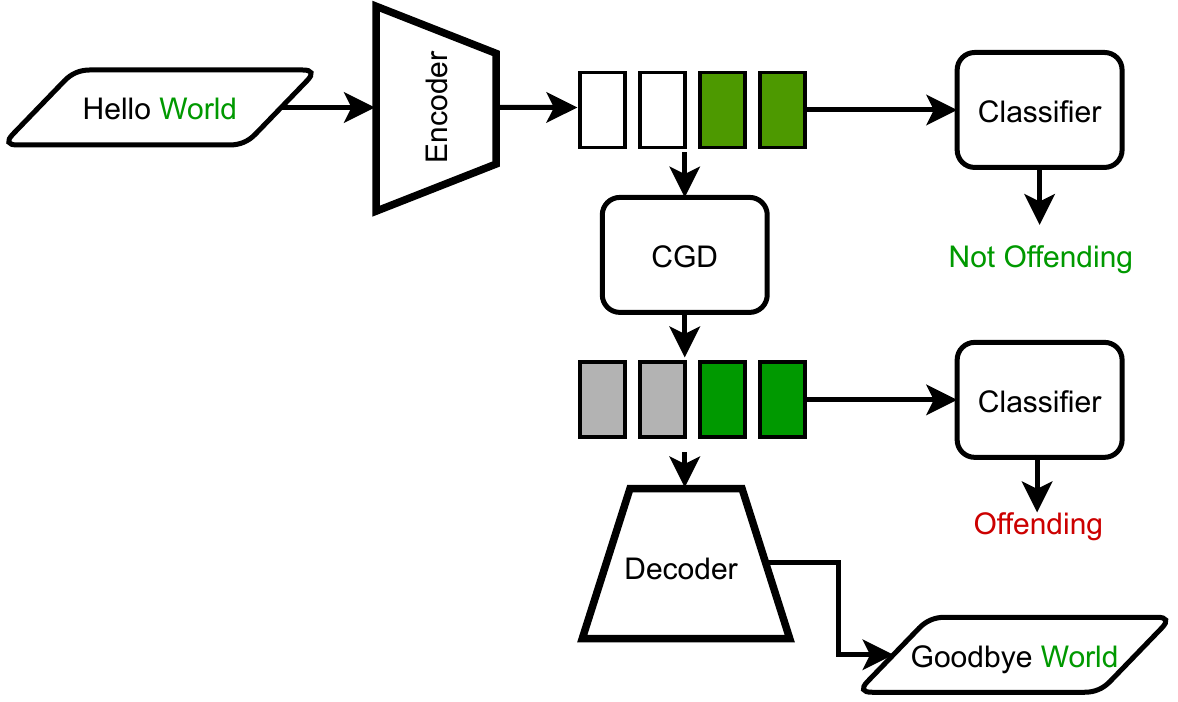}
    \caption{The process of applying CGD in the NLP domain.}
    \label{fig:nlp}
\end{figure}

\noindent \textbf{Experiment Results:} \label{sec:results} Table~\ref{tab:all_results} shows the results aggregated over all benchmark suites for each constraint on the PAR2 score. 

Overall, \toolname was found to have the lowest (i.e., the best) PAR2 score among all the tools on the disguised and flow adversarial constraints, with 309.13\% and over 1500\% improvement over the next best tool, respectively (See Table~\ref{tab:all_results}). Moreover, \toolname was comparable to the AutoAttack tool which was observed to have the best score on $\ell_p$ ball-based adversarial constraints. In the $\ell_p$ ball-based adversarial constraint experiments, B\&B and AutoAttack resulted in the best performance  because they are able to circumvent adversarial training, and so they performed significantly better than other methods on DNN models in the benchmark that were adversarially trained. Even so, \toolname performs comparably to AutoAttack (See Table~\ref{tab:all_results}).

As shown in our experiments, most of the other tools can not find adversarial examples using the other two constraints. While the testing tools were observed to perform well on the $\ell_p$ ball-based adversarial experiments, they were unable to construct inputs satisfying the other two constraints. By contrast, verification/program analysis and reachability tools were able to test for the $\ell_p$ ball-based and disguised constraints, and produced good results on the smaller VNN-LIB suite. However, they failed to scale to larger models and did not support the flow constraint experiments. Moreover, \toolname has a much better performance over other GD-based methods (e.g., FGSM) precisely due to the specification awareness of CGD, where the gradient-based search allows us to satisfy constraints. Finally, fuzzing and GA tools showed consistently poor PAR2 results on all three constraints. 

\subsection{Adversarial Robustness over the NLP Domain} \label{sec:experiment_nlp}

\begin{table}[t]
    \centering
    \begin{tabular}{|l|r|r|}
    \hline
    \textbf{TAV Tools} & \textbf{PAR2 (sec)} & \textbf{ASR} \\ \hline 
    \toolname & \textbf{32,604.40}  & 73\% \\ \hline
    pso & 33,343.85 & \textbf{ 74\%} \\ \hline
    hotflip & 41,440.28 & 67\% \\ \hline
    pruthi & 44,494.51 & 64\% \\ \hline
    textfooler & 51,617.14 & 59\% \\ \hline
    deepwordbug & 55,221.63 & 56\% \\ \hline
    \end{tabular}
    \vspace{0.1cm}
    \caption{Results of the NLP experiment. ASR stands for Attack Success Rate.}
    \label{tab:nlp_results}
\end{table}

In order to allow \toolname (and other GD tools) to test NLP models~\footnote{Note that tools that participate in VNNCOMP are unfortunately not set up to accept NLP input properties.}, we propose the following process (refer to Figure~\ref{fig:nlp}). First, to ensure that the perturbation preserves the syntax and semantics of the original input, we add constraints so that only certain parts of the input are modified. Specifically, our constraints are crafted to perturb named entities only. A named entity is a proper noun in linguistics (i.e., name of a person, location, etc.). In many cases, perturbing named entities does not affect the syntax or semantics of a text~\cite{namedentity}. Named Entity Recognition is a popular NLP task that is performed well by state-of-the-art NLP models. As a pre-processing step, we identify all the named entities in an input and add constraints requiring the adversarial and original input to have minimal change in the subset of features that are not named entities. We then project the discrete text input to a continuous domain using an encoder model ($Enc$). Here, we use fixed-length inputs, adding extra padding at the end of each input. This is done so that we can automatically identify the parts of the continuous embedding region that correspond to the input, so we know the location of the named entities. We then apply the gradient-based CGD attack, ensuring that only the named-entity regions are perturbed, converting the resulting adversarial example back to the discrete domain using a decoder model ($Dec$). Note that we require that $Dec(Enc(x)) = x$ so that the conversion between discrete text to continuous values (and vice versa) does not cause any modifications to the input. In practice, this is not always true, and so we discard any adversarial inputs that are not similar to the original in the non-named entity region.

In order to evaluate \toolname on this domain, we compared it against 5 state-of-the-art NLP adversarial attack tools implemented via the TextAttack framework, using the TweetEval benchmark. The target model was trained to semantically classify whether a given tweet is offensive using the continuous encoding of the input. We sampled 100 inputs and provided all tools with the task of finding adversarial examples and compared the PAR2 score on this suite. 

\noindent {\bf Experiment Results: } Table~\ref{tab:nlp_results} contains the results of this experiment. Once again, \toolname resulted in the best performance, with an improvement in PAR2 score of 2.3\% over the next best tool (pso). Note that pso generated slightly more adversarial examples successfully (74\%) compared to \toolname (73\%). However, our tool was faster in producing adversarial examples and thus, resulted in a better PAR2 score, even though it is not purpose-built for this domain.

\subsection{VNN-COMP 2021 and 2022} \label{sec:vnn_res}

\begin{table}[t]
\begin{center}
\caption{VNN-COMP 2021 Results} \label{tab:score2021}
{\setlength{\tabcolsep}{2pt}
\begin{tabular}{|c|l|c|}
\hline
\textbf{TAV Tools} & \textbf{PAR2 (sec)}\\
\hline
$\alpha$,$\beta$-CROWN & 7,373 \\\hline
Verinet & 8,210 \\\hline
ERAN & 9,770 \\\hline
oval & 10,924 \\\hline
CGDTest & \textbf{11,009} \\\hline
Marabou & 12,851 \\\hline
Debona & 14,222 \\\hline
venus2 & 15,836 \\\hline
Nnenum & 16,095 \\\hline
nnv & 21,352 \\\hline
NV.jl & 21,985 \\\hline
RPM & 22,372 \\\hline
DNNF & 22,849 \\\hline

\end{tabular}
}
\end{center}
\end{table}

We did not officially participate in VNN-COMP 2021, but we conducted extensive experiments comparing \toolname on the VNN-COMP 2021 benchmarks and AWS tool instances provided by the competition organizers. The  tools that participated in VNN-COMP 2021 are - \textbf{Marabou}~\cite{katz2019marabou}, \textbf{VeriNet}~\cite{Henriksen+21}, \textbf{ERAN}~\cite{muller2022ml}, \textbf{$\boldsymbol{\alpha}$,$\boldsymbol{\beta}$-CROWN}~\cite{zhang2022general}, \textbf{DNNF}~\cite{9402125}, \textbf{NNV}~\cite{tran2020nnv}, \textbf{OVAL}~\cite{de2021improved}, \textbf{RPM}~\cite{vincent2021reachable}, \textbf{NV.jl}~\cite{liu2019neuralverification}, \textbf{Venus}~\cite{kouvaros2021towards}, \textbf{Debona}~\cite{brix2020debona}, and \textbf{nnenum}~\cite{bak2021nnenum}. CGDTest stood fifth overall on supported 2021 benchmarks and stood second on acasxu benchmark, and third on marabou-cifar10 and verivital benchmarks. The overall results are tabulated in Table \ref{tab:score2021}.

\begin{table}[t]
\begin{center}
\caption{VNN-COMP 2022 Results} \label{tab:score2022}
{\setlength{\tabcolsep}{2pt}
\begin{tabular}{|c|l|c|}
\hline
\textbf{Rank} & \textbf{TAV Tools} & \textbf{Score}\\
\hline
1 & $\alpha$,$\beta$-CROWN & 1274.9 \\\hline
2 & \textsc{MN-BaB} & 1017.5 \\\hline
3 & Verinet & 892.4 \\\hline
4 & Nnenum & 534.0 \\\hline
5 & CGDTest & \textbf{408.4} \\\hline
6 & Peregrinn & 399.0 \\\hline
7 & Marabou & 372.2 \\\hline
8 & Debona & 222.9 \\\hline
9 & Fastbatllnn & 100.0 \\\hline
10 & Verapak & 98.2 \\\hline
11 & Averinn & 29.1 \\\hline

\end{tabular}
}
\end{center}
\end{table}

We did submit CGDTest to VNN-COMP 2022, where we stood fifth overall among the 11 participating tools. More importantly, CGDTest stood second for three benchmark categories (cifar100-tinyimagenet-resnet, cifar-biasfield, and sri-resnet-a). Note that CGDTest was the only incomplete tool participating in the competition. For every instance in the benchmark, the organizers scored based on the tool's accuracy and relative time taken. The final score is the sum of the normalized score of every benchmark. Table~\ref{tab:score2022} contains the final score for all the participating tools. For more details about the scoring, please refer to Section 2 of the VNN-COMP 2022 report~\cite{vnncomp2022report}.  For the competition, we implemented our tool in Python 3.8 using Pytorch 1.11.0. To allow for comparability of results, the organizers evaluated all the tools on equal-cost hardware using Amazon Web Services (AWS). One of the permitted instances - m5.16xlarge - with 64 vCPUs and 256 GB RAM was used to run our tool. For more details, readers can refer to the VNN-COMP 2022 report~\cite{vnncomp2022report}.

The other tools that participated in the VNN-COMP 2022 are - \textbf{$\boldsymbol{\alpha}$,$\boldsymbol{\beta}$-CROWN} ~\cite{zhang2022general}, \textbf{AveriNN}~\cite{prabhakar2019abstraction}, \textbf{Debona}~\cite{brix2020debona}, \textbf{FastBATLLNN}~\cite{FerlezKS22}, \textbf{Marabou}~\cite{katz2019marabou}, \textbf{\textsc{MN-BaB}}~\cite{mnbab}, \textbf{nnenum}~\cite{bak2021nnenum}, \textbf{PeregriNN}~\cite{khedr2021peregrinn}, \textbf{VeraPak}~\cite{Smith2021}, and \textbf{VeriNet}~ \cite{Henriksen+21}.


\section{Related Work} \label{sec:related}
\noindent {\bf DNN Testing Methods:} Testing methods can be sub-divided into three broad categories, namely, fuzzing, genetic algorithm or GA, and GD methods. GD methods use some form of gradient descent to attempt to find examples that satisfy some properties. Examples include the Fast Gradient Sign Method (FGSM)~\cite{goodfellow2014explaining}, the Carlini \& Wagner (C\&W) attack~\cite{goodfellow2014explaining}, and the Brendel \& Bethge attack~\cite{bandb}. These methods are efficient, extremely generalizable, and require nothing more than access to the model-under-test (and possibly an input to modify). However, they come at the expense of both expressibility and guarantees. On the other hand, \toolname is specification-aware which allows it to generate inputs that can satisfy complex constraints. Another recent tool is DLFuzz~\cite{fuzzy}, which utilizes neuron coverage statistics to guide fuzz in the direction of greater neuron coverage to find adversarial examples. DLFuzz has difficulty scaling to industrial-sized networks and is limited to searching for adversarial examples. By contrast, \toolname scores highly on scalability, expressibility and generality. Note that the class of constraints expressible via CGD is much larger than that in other gradient-based methods, allowing a CGD-based approach to be applied to situations with more complex input constraints.

\noindent {\bf Verification/Analysis Methods:} These methods encode the model-under-test symbolically and represent properties as formulas or constraints, with the aim of providing a certificate of guarantee (i.e., no violations of properties) or produce a counter-example. There are several tools, such as the ones by MIP Verify~\cite{tjeng2017evaluating}, that encode DNNs and properties as MILP problems and use corresponding solvers to prove or disprove the existence of property-violating examples. They often restrict DNN activation functions and architectures to be piecewise-linear DNNs, which is not the case for \toolname. On the other hand, tools like Marabou~\cite{katz2019marabou} are simplex-based methods that can handle arbitrary piecewise linear activation functions. By contrast to these tools, \toolname is efficient and scales to large industrial-sized DNNs. Further, \toolname also scores highly on generality due to using GD at the expense of guarantees and, is comparable to verification tools in terms of expressibility. 

\noindent {\bf Reachability Methods:} These methods define an allowable input region and propagate that region through the network, resulting in an output domain that encapsulates any possible reachable output given one of the input values. Via this method, one can verify robustness by limiting the input region to a shape around an example (such as a perturbation bound) and verifying that the reachable outputs do not include incorrect classifications. Examples of such tools include ERAN~\cite{muller2022ml} and NNV~\cite{tran2020nnv}. Unfortunately, in contrast to \toolname, these methods do not scale well and score low on generality.  

\noindent {\bf Learning via Constraint Loss:} To the best of our knowledge~\cite{psl} were the first to introduce \textit{probabilistic soft logic}, which allowed for encoding logical constraints as loss functions, that are mostly continuous and used solvers to infer the most probable explanation (MPE). \cite{dl2} then introduced DL2, a tool that allows for certain logical constraints to be converted into differential (almost) everywhere loss functions. The resulting constraint loss function was used by these tools to train a DNN in a manner that minimizes constraint violations by updating DNN parameters so as to minimize the constraint loss. By contrast, our CGD method differs from the above-mentioned methods in the following ways: first, we combine constraint loss with a modified GD-based method (specifically FGSM~\cite{goodfellow2014explaining}). Further, our method is aimed at generating DNN inputs that satisfy certain constraints, as opposed to DL2 that uses constraint-loss-based GD method to train DNNs.

\noindent {\bf DNN Adversarial Attacks in the NLP Domain:} Unlike in the vision domain, gradient-based adversarial attacks do not typically work in the natural language processing domain because the inputs are discrete, and they possess inherent syntactic and semantic restrictions. Further, research has shown that even state-of-the-art language models used in the industry are vulnerable to adversarial attacks. Currently, researchers have developed DNN TAV tools specific to this domain that use combinatorial search in order to find adversarial inputs that obey the required constraints. Most search procedures either modify parts of the input (such as flipping a character or swap specific parts of the input with similar words. TextAttack~\cite{textattack} is a popular framework that implements several state-of-the-art NLP adversarial attacks - which we compare against, and show that \toolname outperforms them. That is, \toolname is a very general tool, and yet it is able to perform as well or better against state-of-the-art specially-designed domain-specific testing tools.

\noindent {\bf Testing for Individual Fairness:}
\cite{individual_fairness_old} formulated testing for individual fairness as a form of robustness to perturbations of sensitive inputs of a model. However, this work looks at training fair models instead of testing for them. By contrast, the work of \cite{individual_fairness} introduced a method to find unfair points using a gradient-based search that perturbs sensitive features of the input. By contrast, our tool is a more general method that can be used to test for various constraints - including this version of individual fairness.

\noindent {\bf CGD vs PGD:}
Projected Gradient Descent (PGD)~\cite{madry2017towards} minimizes the loss function by moving in the direction of the negative gradient at each step and then ``projecting" onto the feasible set defined in the constraint. Note that this projection is performed after gradient descent has been completed. By contrast, CGD converts the set of constraints into a differentiable form, simultaneously computes constraint and misclassification loss, and applies gradient descent directly. 

\section{Conclusions and Future Work} \label{sec:conclusion}

To address the reliability and security problems associated with DNNs, we propose a new testing algorithm called the Constrained Gradient Descent (CGD) method, which is a gradient-descent method whose loss function takes into account user-specified logical constraints. Via an extensive empirical evaluation of \toolname against \numoftool other state-of-the-art methods using $\ell_p$ ball-based and other richer classes of adversarial examples such as flow constraints, we show that \toolname results in an improvement in PAR2 score of up to 1500\% over the next best method in the vision domain. We also show the efficacy of \toolname when applied to the domain of NLP by projecting inputs to a continuous embedding and adding constraints to respect the syntax and semantic restrictions inherent to NLP. In this setting, our experiments comparing \toolname with 5 state-of-the-art NLP TAV tools show that our method results in an improvement in PAR2 score of 2.2\% over the next best special-purpose testing method. We also stood fifth among the 11 participating tools in VNN-COMP 2022, even though \toolname was the only incomplete tool in the competition. 

Our extensive experiments showcase the power of our tool \toolname as compared to leading-edge TAV methods, in that it scales well on large real-world models with over 138 million parameters (scalability), allows users to specify a wide variety of DNN properties over multiple domains (expressibility), and places no restrictions on the kinds of DNNs it can analyze (generality). All other tools we compared against fared worse on at least one of these vectors. For example, verification tools that dominate in the VNNCOMP competition are not equipped to handle flow constraints or the NLP domain, and further cannot handle many common DNN architectures because of the difficulty involved in converting them into a symbolic form. Further, fuzzing or GA tools are narrowly tailored for adversarial inputs in the vision domain. Similarly, NLP testing tools are applicable only to NLP models. In the future, we plan to extend our tool to other domains such as safety-critical settings, as well as use it as a component in a complete verification tool. 

\bibliographystyle{named}
\bibliography{cgdtest}

\clearpage
\appendix
\section{Terminology and Constraint Translation}\label{app:conversion}

In this section, we provide information on the terminology and process of converting constraints to differentiable loss functions. We refer readers to~\cite{psl,dl2} for more details.

\subsection{Terminology}
The terminology used in this work is consistent with and is a subset of the language used in boolean logic. This consists of different terms that are connected using several symbols. We use the following symbols in our work:

\begin{enumerate}
    \item $\varphi$: Represents a formula or a sub-formula. Note that this can be of arbitrary length and is recursively defined using other symbols (including itself).
    \item $\wedge$: boolean AND
    \item $\neg$: NOT
    \item $\vee$ boolean OR
    \item comparison: This refers to a term that consists of two $t$ terms that are connected using $\wedge$ or $\vee$.
    \item $t$: A term that is either a differentiable function (such as $\ell_p$ distance) or some constant.
\end{enumerate}

\subsection{Constraint Relaxation}
In traditional logic, each constraint (or rule) is mapped to either a 1 (true) or 0 (false) truth value. It is hard to capture uncertainty in this setting - a trait very important to many AI problems. Further, solving in traditional logic can be computationally intractable. By contrast, probabilistic soft logic (PSL)~\cite{psl} - a framework for collective, probabilistic reasoning in relational domains - uses soft truth values in the interval [0,1]. This allows inference in this setting to be a continuous optimization problem, which can be solved efficiently. Additionally, one can encode similarity functions and other relationships as constraints in this domain.

In order to determine the degree that a constraint is satisfied, PSL uses the \textit{Lukasiewicz t-norm} and its corresponding \textit{co-norm} as the relaxation of logical connectives. These relaxations, as well as the truth values of the constraints, are designed to be exact at the extremes (`completely true' or `completely false') - and more importantly - provide a consistent mapping for values in between. This allows us to calculate smooth gradient information. Examples of these relaxations are provided below. 

\subsection{The Language of Constraints used in \toolname} \label{sec:conversion}
Similar to~\cite{dl2}, the language of constraints in our setting consists of boolean combinations of comparisons between terms, where each term is either a constant or a differentiable function. More formally, the language of constraints can be given as follows:

\begin{itemize}
    \item $\varphi \Rightarrow \varphi \wedge \varphi \; | \; \varphi \vee \varphi \; | \; \neg \varphi \;| \;$ comparison
    \item comparison $\Rightarrow t \leq t \; | \; t = t \; | \; t \neq t \; | \; t < t$
    \item $t \Rightarrow $ constant $|$ differentiable function
\end{itemize}

Each term $t$ can be a constant or a real-valued differentiable function that is defined over inputs, parameters (such as intermediate layer outputs) and outputs of the DNN. For our implementation, we included a number of popular differentiable functions (e.g. $\ell_{\infty}$ distance), which can be extended to more functions, as long as they are differentiable. 

\subsection{Translating Constraints to Differentiable Loss}

The resulting formula $\varphi$ is converted using a recursive descent parsing algorithm to a differentiable loss function. The main idea behind this algorithm is to map boolean connectives and comparison terms to an approximate relaxation using \textit{Lukasiewicz t-norm} and its corresponding \textit{co-norm}. The process is given below in detail.

The comparison terms are converted as follows:
\begin{enumerate}
    \item $\mathcal{L}(t\leq t') := max(t-t',0),$
    \item $\mathcal{L}(t\neq t') := \epsilon*[t=t'],$
    \item $\mathcal{L}(t = t') := \mathcal{L}(t\leq t' \wedge t' \leq t),$
    \item $\mathcal{L}(t < t') := \mathcal{L}(t\leq t' \wedge t \neq t').$
\end{enumerate}

The boolean combination of constraints are converted as follows:
\begin{enumerate}
        \item $L(\varphi' \wedge \varphi'') := L(\varphi') + L(\varphi''),$
        \item $L(\varphi' \vee \varphi'') := L(\varphi') * L(\varphi'')$.
\end{enumerate}

If $\varphi$ is a negation $\neg\varphi'$, it is rewritten to a logically equivalent constraint before translation to loss as follows:
\begin{enumerate}
    \item $\mathcal{L}(\neg(t = t')) := \mathcal{L}(t\neq t'),$
    \item $\mathcal{L}(\neg(t \leq t')) := \mathcal{L}(t' < t),$
    \item $\mathcal{L}(\neg(t \neq t')) := \mathcal{L}(t = t'),$
    \item $\mathcal{L}(\neg(t < t')) := \mathcal{L}(t'\leq t),$
    \item $\mathcal{L}(\neg(\varphi' \wedge \varphi'')) := \mathcal{L}(\neg\varphi' \vee \neg\varphi''),$
    \item $\mathcal{L}(\neg(\varphi' \vee \varphi'')) := \mathcal{L}(\neg\varphi' \wedge \neg\varphi''),$
    \item $\mathcal{L}(\neg(\neg \varphi')) := \mathcal{L}(\varphi').$
\end{enumerate}

This process produces a function with the following properties:
\begin{enumerate}
    \item The value is non-negative, and the greater the number of violations, the higher the value produced.
    \item A value of 0 corresponds to satisfying all constraints fully.
    \item It is sub-differentiable (i.e., differentiable almost everywhere).
\end{enumerate}

Because of these properties, we can use a GD algorithm that modifies an example along the gradient of reduced constraint loss, as that corresponds to finding an input that better satisfies these constraints. Alternatively, we can also use this in a genetic algorithm as a fitness function, with the goal of finding candidates that minimize this value.




\section{Constraints} \label{app:constraints}

In this section, we provide additional information on the constraints used in our experiments. We use five constraints in our experiments:

\subsection{$\ell_p$ ball-based Adversarial Constraints}

A $\ell_p$ ball-based adversarial example uses $\ell_\infty$ in order to measure similarity and is defined as:

\begin{center}
$(\|x-x'\|_\infty \leq \delta) \wedge (argmax(M(x)) \neq argmax(M(x')))$
\end{center}

Once the constraints are converted to the differentiable constraint loss (using the translations provided in Appendix~\ref{app:conversion}), we get:

\begin{center}
$max(\|x-x'\|_\infty-\delta,0)+\eta*[argmax(M(x))=argmax(M(x'))]$.
\end{center}

Here, $\eta > 0$ is a parameter that controls the loss value for the inequality, and $[argmax(M(x))=argmax(M(x'))]$ is an indicator function that returns 1 if $M$ classifies $x$ and $x'$ using the same label, or 0 otherwise. As can be seen, the resulting function results in 0 only if $x'$ is an adversarial example as given by definition, or is greater than 0 otherwise. This is the constraint loss that is given in line 2 of Algorithm~\ref{alg:cgd}. Our experiments use a $\delta$ value of 0.02, and $\eta$ value of 0.5.

\subsection{Disguised Adversarial Constraints} \label{app:disguised}

A disguised adversarial constraint requires the original label to be the second-highest label of $M$. Here, $x'$ is in some notion still similar to $x$ as $M$ has fairly high confidence in $y$ (since it is the second highest label). In the previous definition using $\ell_p$ ball-based adversarial constraint, we require the new label on $x'$ be anything \textit{but} $y$, while here we require $y$ to be the \textit{second highest} label. They are also subtly different from targeted adversarial examples, whose goal is to cause the $M$ to classify $x'$ as a different label $y' \neq y$.

Consider the task of handwritten digit recognition. The digits 1 and 7 look similar for certain handwritings. So, while performing the classification task, we would be more interested in the cases where the number 1 is misclassified as 7 or vice-versa.

Similarly, consider the following scenario of an image recognition system deployed in an airport aimed at identifying a bad actor from their picture. When given an undisguised picture, the model would correctly identify the bad actor. However, chances are that the bad actor would be disguised. In this scenario, the disguised image may be classified in top-2, but not necessarily the highest. The disguised adversarial examples would attack the model and force it to misclassify the top-2 inputs. 

In order to ensure this, we add constraints such that $M$ does not classify $x'$ as $y$, but $y$ is greater than all labels but one. In other words: 

\begin{center}
$argmax(M(x') \neq y \wedge (\bigvee_{y'\in Y-y} \bigwedge_{y''\in Y-\{y,y'\}} argmax(M(x')) > y'')$
\end{center}

In other words, we first set the constraint that $M$ labels $x'$ as something other than $y$. Further, we add constraints so that the label of $M$ on $x'$ is greater than all but one label. We omit the differentiable function for brevity.

\subsection{Flow Constraints}

Flow constraint utilizes spatial transformations (flow) to find adversarial examples with minimal spatial transformation. It is given as: 

\begin{center}
$(L_{Flow}(x,x')\leq\delta)\wedge(argmax(M(x))\neq argmax(M(x')))$, 
\end{center}

where $L_{Flow}$ implements the flow loss described in Xiao et al.~\cite{xiao2018spatially}. 

Similar to $\ell_p$ ball-based adversarial constraint, this is translated as:

\begin{center}
$max(L_{flow}(x,x')-\delta,0)+\eta*[argmax(M(x))=argmax(M(x'))]$
\end{center}

Note that previous testing methods simply optimize to increase misclassification scores to find an adversarial example and ensure similarity to the original input via small incremental updates. This works well most of the time, but not always since we are not explicitly looking for similar inputs. By contrast, CGD directly optimizes based on both increased misclassification scores as well as similarity.

\subsection{NLP Adversarial Constraints}

In the case of NLP, constraints are placed on the non-named entity regions. As a pre-processing step, each word is checked if it is a named entity using the \texttt{nltk} package in python. The text is then projected to a fixed-length continuous domain, and the region in the continuous space that corresponds to the non-named entity tokens is identified. Note that because we use a fixed-length vector for the projection, we add padding to the text inputs that are smaller in length. 

Once we identify the non-named entity regions, we simply add constraints so that all features within these regions are the same between the original and adversarial inputs. Given region $R$ that denotes all features that are non-named entities, the adversarial constraints in NLP are given as: 

\begin{center}
$(\bigwedge\limits_{r=1}^R x_r = x'_r) \wedge (argmax(M(x))\neq argmax(M(x'))) $
\end{center}

The corresponding differentiable function is: 

\begin{center}
$(\sum_{r=1}^{R} (max(x_r-x'_r, 0) + max(x'_r-x_r, 0))) + \eta*[argmax(M(x))=argmax(M(x'))]$
\end{center}

\subsection{Fairness Constraints}

For our fairness constraints, we use the distance metric learned from data in~\cite{individual_fairness}. Logistic regression classifiers are designed to predict the protected variables, and the distance $d^{2} = ||(I - \prod_{\mathcal{H}})x - x'||_2^{2}$, where $\mathcal{H}$ is the sensitive subspace learned by the regression models and $\prod_{\mathcal{H}}$ is the projection of it.

Once we have this distance metric, finding unfair points equates to:

\begin{center}
$argmax(M(x)) \neq argmax(M(x')) \wedge ||(I - \prod_{\mathcal{H}})x - x'||_2^{2}$
\end{center}

The corresponding differentiable function is:

\begin{center}
$\eta*[argmax(M(x))=argmax(M(x'))] + max(||(I - \prod_{\mathcal{H}})x - x'||_2^{2} - \delta, 0)$
\end{center}

\subsection{Motivation}

Our motivation for using the latter four constraints is to show a great weakness of existing DNN TAV tools that are ALL unfortunately based on $\ell_p$-ball verification. In other words, if a DNN is robust against perturbed images within an $\ell_p$-ball of the original image, then such DNNs would be certified as resistant to adversarial attacks by today's verification tools (assuming that these verification tools scale). Unfortunately, such a certificate does not mean much because, as previously mentioned, an attacker can still attack such DNNs via these other adversarial constraints. Further, DNN TAV tools are currently tailored toward a specific domain. This is because different domains, such as NLP, pose additional constraints and have discrete inputs. Our method of projecting the input to a continuous encoding and adding constraints to respect the restrictions inherent to the domain can be potentially applied to other domains, allowing for a DNN TAV tool that works in multiple domains.

\end{document}